\documentclass{llncs}

\usepackage{url}
\usepackage[hidelinks]{hyperref}
\usepackage[utf8]{inputenc}
\usepackage{amsmath}
\usepackage{booktabs}
\usepackage{xfrac}

\usepackage{amssymb}
\usepackage{array}   

\newcommand{\lsem}{\mbox{$\lbrack\!\lbrack$}}
\newcommand{\rsem}{\mbox{$\rbrack\!\rbrack$}}
\newcommand\mdoubleplus{\mathbin{+\mkern-10mu+}}

\newcolumntype{L}{>{$}l<{$}} 
\newcolumntype{R}{>{$}r<{$}} 
\newcolumntype{C}{>{$}c<{$}} 

\usepackage{caption}

\newif\ifarxiv
\newif\ifincludeproof

\arxivtrue

\ifarxiv
        \includeprooftrue
\else
        
\fi

\begin{document}

\title{Lexicographic Logic: a Many-valued Logic for Preference Representation}

\author{%
Angelos Charalambidis\orcidID{0000-0001-7437-410X}
\and Giorgos Papadimitriou
\and Panos Rondogiannis
\and Antonis Troumpoukis
}
\institute{University of Athens \\
\email{\{acharala,gspapajim,prondo,antru\}@di.uoa.gr}
}

\maketitle
\pagestyle{plain}

\vspace{-0.5cm}

\begin{abstract}
Logical formalisms provide a natural and concise means for specifying and
reasoning about preferences. In this paper, we propose {\em lexicographic
logic}, an extension of classical propositional logic that can express a variety
of preferences, most notably lexicographic ones. The proposed logic supports a
simple new connective whose semantics can be defined in terms of finite lists of
truth values. We demonstrate that, despite the well-known theoretical
limitations that pose barriers to the quantitative representation of
lexicographic preferences, there exists a subset of the rational numbers over
which the proposed new connective can be naturally defined.  Lexicographic logic
can be used to define in a simple way some well-known preferential operators,
like ``$A$ and if possible $B$'', and ``$A$ or failing that $B$''. Moreover,
many other hierarchical preferential operators can be defined using a systematic
approach. We argue that the new logic is an effective formalism for ranking
query results according to the satisfaction level of user preferences.
\end{abstract}

\section{Introduction}
\label{intro}
During the past three decades, many formalisms have been developed for
representing preferences, both in artificial intelligence~\cite{DomshlakHKP11}
as-well-as in databases~\cite{StefanidisKP11}. Of particular interest are the
logical approaches in which the specification of preferences is performed using
operators that implicitly manipulate the underlying preference
values~\cite{BrewkaBB04,BrewkaNT08,AW05,Lang09,DuboisP13,RondogiannisT15,CharalambidisRT18}.
Such formalisms are usually declarative, concise, and easy to understand and
reason about.

In this paper, we develop {\em lexicographic logic}, a simple extension of
classical propositional logic that can express a variety of preferences, most
notably lexicographic ones. The proposed logic adds only one formation rule to
the syntax of propositional logic: if $\phi_1$ and $\phi_2$ are formulas, then
so is $(\phi_1 \gg \phi_2)$. The formula $(\phi_1 \gg \phi_2)$ can be read
``{\em $\phi_1$ has a lexicographic priority over $\phi_2$}'' (a notion that
will be explained in detail in Section~\ref{motivation}). The semantics of
``$\gg$'' can be defined in terms of finite lists of truth values of a
three-valued logic. Actually, we demonstrate that such lists have a natural
mapping to rational numbers in the interval $[-1,1]$, and the meaning of
``$\gg$'' can then also be understood as a function that maps pairs of rational
numbers to rational numbers. This interpretation of ``$\gg$'' allows the
evaluation of lexicographic logic formulas using standard floating-point
arithmetic. Apart from its simplicity, an advantage of lexicographic logic is
that it can be used to represent concisely well-known preferential operators
as-well-as to define new ones. The main contributions of the paper can be
summarized as follows:
\begin{itemize}
\item We define a novel, non-classical, propositional logic for expressing
      preferences. The primary connective of this logic expresses lexicographic
      priority, which is known to be non-trivial to specify from a quantitative
      point of view~\cite{Fishburn99}. We demonstrate that the semantics of this
      new connective can be specified quantitatively as a function over rational
      numbers. The main theorem of the paper
      (Theorem~\ref{lexicographic-theorem}) asserts that ``$\gg$'' ensures
      strict monotonicity with respect to lexicographic comparison.

\item We present the properties of lexicographic logic and investigate its
      connections with alternative operators that have been proposed in the
      literature. We demonstrate that some well-known and useful such operators
      can be succinctly represented using simple formulas of lexicographic
      logic. Additionally, we propose a systematic approach for representing new
      {\em hierarchical} preferential operators as derived formulas of
      lexicographic logic.
\end{itemize}
The rest of the paper is organized as follows. Section~\ref{motivation} provides
the motivation and intuition behind the proposed approach.
Section~\ref{syntax-semantics} presents the syntax and semantics of
lexicographic logic. Section~\ref{properties} establishes properties of the
proposed logic, and Section~\ref{numerical} demonstrates that the semantics of
the logic can alternatively be defined in a numerical way. In
Section~\ref{alternative} it is argued that formulas of the proposed logic can
be used to define a variety of hierarchical preferential operators.
Section~\ref{related-work} presents related work on preference representation
formalisms and Section~\ref{conclusions} concludes the paper giving pointers to
future work.

\section{Motivation and Intuition}
\label{motivation}
In this paper, we contribute to the logical specification of preferences by
introducing {\em lexicographic logic}, an extension of propositional logic for
describing lexicographic (as-well-as many other) preferences. More specifically,
we add to the syntax of propositional logic the new connective ``$\gg$''.  The
formula $(\phi_1 \gg \phi_2)$ is read ``{\em $\phi_1$ has a lexicographic
priority over $\phi_2$}''. Intuitively, lexicographic priority can be explained
using levels of preference satisfaction, as follows:
\begin{itemize}
\item if both $\phi_1$ and $\phi_2$ are true, then we are completely satisfied;

\item if $\phi_1$ is true and $\phi_2$ is false, then we are satisfied but not
      entirely;

\item if $\phi_1$ is false and $\phi_2$ is true, we are dissatisfied, but not
      strongly so;

\item finally, if both $\phi_1$ and $\phi_2$ are false, then we are completely
      dissatisfied.
\end{itemize}
The following example motivates the above ideas.
\begin{example}\label{example1}
Consider the specification of our preferences for buying a new car. The formula
${\tt (electric}\gg{\tt fast)}$ means that if we buy a car that is both electric
and fast, we will be completely satisfied; if we buy one that is electric but
not fast, then we will not be entirely satisfied; if we buy a car that is only
fast, then we will be dissatisfied (but not entirely); and if none of our
preferences is satisfied, then this will be our least preferred state of
affairs.
\end{example}

To specify the formal semantics of $(\phi_1 \gg \phi_2)$, we use a many-valued
logic to express the different levels of preference satisfaction discussed
above. In particular, if $F,T$ denote the classical false and true values, we
expect that:
\[
      (F \gg F) \, <\,  (F \gg T) \, < \, (T \gg F) \, < \, (T \gg T)
\]
where ``$<$'' can be read as ``less preferred than''. This suggests that we
probably need a four-valued logic to express these four truth levels, which we
could call ``false'', ``less false'', ``less true'', and ``true''. However, it
turns out that four truth levels are not enough. Imagine, for example, that we
apply ``$\gg$'' on a pair consisting of the ``false'' and the ``less false''
truth values. This will give a new truth value between ``false'' and ``less
false''. Actually, one can verify that we need an infinity of truth values if we
want to be able to specify formulas with an arbitrary number of occurrences of
``$\gg$''. The question we have to answer is how this set of truth values is
exactly structured and how are its elements ordered.

To understand the difficulties of defining such a set, we need to become
slightly more formal. Let us denote by $\mathbb{V}$ this (yet unknown) set and
let ``$<$'' be the (yet undefined) ordering relation on the elements of
$\mathbb{V}$. Let $u_1,u_2,v_1,v_2 \in \mathbb{V}$ be truth values. We define
the ``lexicographically smaller'' relation $<_L$ on $\mathbb{V} \times
\mathbb{V}$, as follows:
\[
(u_1,u_2) <_{L} (v_1,v_2) \mbox{ iff } (u_1< v_1) \vee ((u_1 = v_1) \wedge (u_2 < v_2))
\]
The meaning of ``$\gg$'' should be a function $f$ that respects the $<_L$
ordering. This means that for all truth values $u_1,u_2,v_1,v_2$ it must hold:
\[
      (u_1,u_2) <_{L} (v_1,v_2) \mbox{ iff } f(u_1,u_2) < f(v_1,v_2)
\]
One could view truth values as real numbers and attempt to define $f$ as a
function $f:(X,X) \rightarrow X$ where $X$ is a subset of the set $\mathbb{R}$
of real numbers. For example, one could view the classical truth values $F$ and
$T$ as the real numbers $-1$ and $1$ respectively, and try to define ``$\gg$''
as a function $f:([-1,1],[-1,1]) \rightarrow [-1,1]$. However, there is a
well-known obstacle in such an approach, which is described by the following
folk theorem (see for example~\cite[p. 363]{Fishburn99} for a proof):
\begin{theorem}\label{folk-theorem}
There does not exist a function $f:(\mathbb{R},\mathbb{R})
\rightarrow\mathbb{R}$ such that
$(u_1,u_2) <_{L} (v_1,v_2) \mbox{ iff } f(u_1,u_2) < f(v_1,v_2)$.
\end{theorem}
The proof of the above theorem can easily transfer to the case where we replace
$\mathbb{R}$ with the closed interval $[-1,1]$. Therefore, we conclude that if
we would like to specify the semantics of the binary operator ``$\gg$'', we
would have to focus on more refined and carefully constructed truth domains than
the set of the real numbers (or intervals of it).

To understand how we bypass the above problem, we return to
Example~\ref{example1}. Our first idea (which we will subsequently slightly
refine) is that we can express the different levels of preferences by using a
truth domain whose elements are {\em lists} of classical truth values $F$ and
$T$. The semantics of the ``$\gg$'' operator can then be defined as the
concatenation of such lists. In our example, if both ${\tt electric}$ and ${\tt
fast}$ are true, we assign to the formula the value $[T,T]$; if ${\tt electric}$
is true and ${\tt fast}$ is false, we assign the value $[T,F]$; if ${\tt
electric}$ is false and ${\tt fast}$ is true, we assign the value $[F,T]$; if
both atoms are false, we assign the value $[F,F]$. Notice that these lists, when
viewed as words compared lexicographically over the alphabet $\{F,T\}$, where
$F<T$, express precisely the four different levels of truth (``false'', ``less
false'', ``less true'', and ``true'') that we desire for this formula, namely:
\[
      [F,F]\, <\, [F,T] \, <\, [T,F]\, <\, [T,T]
\]
In the following, even the classical truth values $F$ and $T$ will be written in
list form, namely $[F]$ and $[T]$. Notice also that we expect $[T,T]$ to be
equal to $[T]$, because this is a situation where all our preferences are
satisfied; similarly, we expect $[F,F]$ to be equal to $[F]$.

There is, however, a further refinement of the above scheme that is required.
The basic complication that needs to be addressed is that the operator ``$\gg$''
should not be associative. This issue is illustrated by the following example.
\begin{example}\label{car-example}
Consider the formulas  ${\tt electric} \gg {\tt (fast \gg blue)}$ and
${\tt (electric} \gg {\tt fast)} \gg {\tt blue}$. We claim that these two
formulas should not be semantically equivalent. To see this, consider a
truth assignment that assigns to ${\tt electric}$ the value $[T]$, to ${\tt
fast}$ the value $[F]$, and to ${\tt blue}$ the value $[F]$. Intuitively
speaking, the first formula evaluates to $[T]\gg [F,F]$ while the second one to
$[T,F]\gg [F]$. Comparing these two values, we see that in the former one, our
primary requirement is fully satisfied (truth value $[T]$), while in the latter
one, our primary requirement is only partially satisfied (truth value $[T,F]$).
In other words, under this truth assignment, the first formula is ``more
satisfied'' than the second one.
\end{example}
The above discussion suggests that the meaning of ``$\gg$'' should not be
associative, and therefore it should not be defined as just list-concatenation
(which is an associative operator). To properly define the semantics of ``$\gg$'',
we will use one extra truth value, namely $0$. By prefixing each concatenation
operation with a $0$, formulas such as the ones that appear in the above example
will be discriminated. As William W. Wadge observed\footnote{Personal communication.},
our use of the zeros inside the lists essentially mimics the {\em Polish notation}
of the parenthesized expressions. In this way, different lists are created for
different parenthesizations of an expression, and this ensures non-associativity.
Moreover, as we will demonstrate shortly (see Theorem~\ref{lexicographic-theorem})
this simple operation ensures preservation of the lexicographic property. More
specifically, we demonstrate that ``$\gg$'', when viewed as a binary infix function,
satisfies the property:
$$(u_1,u_2) <_{L} (v_1,v_2) \mbox{ iff } (u_1 \gg u_2) < (v_1 \gg v_2)$$
for all $u_1,u_2,v_1,v_2 \in \mathbb{V}$. Therefore, our definition of ``$\gg$''
bypasses the restriction posed by Theorem~\ref{folk-theorem}, by relying on a
carefully selected truth domain.

Actually, we demonstrate that the truth domain $\mathbb{V}$ corresponds to a subset of the
set $\mathbb{Q}$ of rational numbers: the three-valued lists of $\mathbb{V}$ can
be viewed as numbers in the {\em balanced-ternary number system}~\cite{knuth1998art2}
and can be transformed into standard decimal rational numbers in the interval
$[-1,1]$. Therefore, lexicographic logic formulas can be easily evaluated using
standard floating-point arithmetic.

\section{\mbox{Lexicographic Logic: Syntax and Semantics}}
\label{syntax-semantics}

In this section, we introduce the syntax and the semantics of lexicographic
logic and motivate it with examples. The syntax of the proposed logic 
extends that of propositional logic with a new formation rule:
\begin{definition}
Let ${\cal A}$ be a set of propositional atoms. The set of well-formed formulas
of lexicographic logic is inductively defined as follows:
\begin{itemize}
\item Every element of ${\cal A}$ is a well-formed formula,

\item If $\phi_1$ and $\phi_2$ are well-formed formulas, then $(\phi_1 \wedge
      \phi_2)$, $(\phi_1 \vee \phi_2)$, $(\neg \phi_1)$, and $(\phi_1 \gg
      \phi_2)$, are well-formed formulas.

%
\end{itemize}
\end{definition}
For simplicity reasons, we will often omit the outermost parentheses from formulas.
Moreover, to simplify the readability of expressions that contain multiple occurrences
of ``$\gg$'', we will assume that this operator associates to the right. So, for
example, $x \gg (y \gg z)$ will be written as $x \gg y \gg z$.
%

%
The truth domain of lexicographic logic is denoted by $\mathbb{V}$, and consists of
lists of the truth values $F$, $0$, and $T$. By overloading notation, we will use the
symbol ``$\gg$'' to also denote an operation on lists that corresponds to the meaning of
the syntactic element ``$\gg$''. More formally:
\begin{definition}\label{concatenation-definition}
Let $u,v$ be lists of the elements $F$, $T$, and $0$. We define:
$$(u \gg v)  =  \left\{ \begin{array}{ll}
                                   \mbox{$[F]$}, & \mbox{if $u  =  v = [F]$}\\
                                   \mbox{$[T]$}, & \mbox{if $u  =  v = [T]$}\\
                                   \mbox{$[0]$} \mdoubleplus u  \mdoubleplus v, & \mbox{otherwise}
                                        \end{array}\right.$$
where $\mdoubleplus$ is the list concatenation operation.
\end{definition}
We can now precisely define the truth domain of lexicographic logic:
\begin{definition}
The set $\mathbb{V}$ of truth values is the set inductively defined as follows:
\begin{itemize}
\item $[F] \in \mathbb{V}$ and $[T] \in \mathbb{V}$.

\item If $u,v \in \mathbb{V}$, then $(u \gg v)\in \mathbb{V}$.
\end{itemize}
\end{definition}
Notice that due to Definition~\ref{concatenation-definition}, lists of the form
$[0,T,T]$ and $[0,F,F]$ are not allowed (because they are considered identical
to $[T]$ and $[F]$ respectively).

Each element of $\mathbb{V}$ represents some degree of ``true'' or ``false''.
To understand whether a given element is true or false, it suffices to look
at its $\mathit{sign}$:
\begin{definition}
For every $v \in \mathbb{V}$ we define $\mathit{sign}(v)$ to be the leftmost non-zero element
of $v$.
\end{definition}
Therefore, $[0,F,0,F,T]$ is a false value while $[0,T,F]$ a true one.

Given an arbitrary element $v \in \mathbb{V}$, we denote with $\overline{v}$ the
list that results from $v$ by inverting each $F$ to $T$ and each $T$ to $F$. For
example, $\overline{[0,F,0,T,F]} = [0,T,0,F,T]$.
\begin{lemma}
For every $v \in \mathbb{V}$, $\overline{v} \in \mathbb{V}$.
\end{lemma}
\ifincludeproof
\begin{proof}
By a straightforward structural induction.
\end{proof}
\fi
The following property of the elements of $\mathbb{V}$ is easy to
establish:
\begin{lemma}
For every $v \in \mathbb{V}$, the length of $v$ is odd.
\end{lemma}
\ifincludeproof
\begin{proof}
By a straightforward structural induction.
\end{proof}
\fi
As it turns out, no element of $\mathbb{V}$ is a proper prefix of another element.
We will use this property in the proof of the main theorem of the paper
(Theorem~\ref{lexicographic-theorem}).
\begin{lemma}\label{prefix_lemma}
Let $u,v\in \mathbb{V}$. If $u$ is a prefix of $v$, then $u=v$.
\end{lemma}
\ifincludeproof
\begin{proof}
By induction on the length of $u$. If $|u|=1$ then, $u = [F]$ or $u=[T]$, and it
must also be $v = [F]$ or $v=[T]$, respectively. Assume the lemma holds for all
elements of $\mathbb{V}$ of length $\leq k$.  We demonstrate the statement for
elements of length $k+2$. Assume $u = [0]\mdoubleplus u_1 \mdoubleplus u_2$ and
$v = [0]\mdoubleplus v_1 \mdoubleplus v_2$. Since $u$ is a prefix of $v$, $u_1
\mdoubleplus u_2$ is a prefix of $v_1 \mdoubleplus v_2$, which implies that
either $u_1$ is a prefix of $v_1$ or $v_1$ is a prefix of $u_1$. By the
induction hypothesis, $u_1=v_1$. This implies that $u_2=v_2$ and therefore
$u=v$.
\end{proof}
\fi
By assuming the ordering $F < 0 < T$, any two elements of $\mathbb{V}$ can be
compared lexicographically:
\begin{definition}\label{lexicographic-ordering}
Let $u,v\in \mathbb{V}$ and assume $u=[u_1,\ldots,u_k]$ and
$v=[v_1,\ldots,v_m]$. We will write $u < v$ if there exists $1 \leq r \leq
\min\{k, m\}$ such that
$u_1=v_1,\ldots,u_{r-1}=v_{r-1}$ and $u_r < v_r$.
We will write $u \leq v$ if either $u = v$ or $u < v$.
\end{definition}
Notice that, due to Lemma~\ref{prefix_lemma}, for any $u,v \in \mathbb{V}$,
it will either be $u \leq v$ or $v \leq u$. It is easy to see that
the lexicographic ordering $\leq$ of Definition~\ref{lexicographic-ordering}
is a total ordering on $\mathbb{V}$.

The notion of truth assignment can be generalized as follows:
\begin{definition}
A {\em truth assignment} is a function from the set ${\cal A}$ of propositional
atoms to the set $\mathbb{V}$ of truth values.
\end{definition}
The meaning of a formula is an element of $\mathbb{V}$. More specifically, the
semantics of lexicographic logic is defined as follows:
\begin{definition}\label{semantics}
Let $\phi_1$, $\phi_2$, and $\phi$ be formulas and let $I$ be a truth
assignment. The meaning of a formula with respect to $I$ is recursively
defined as follows:
\begin{itemize}

\item $\lsem \mathsf{p} \rsem (I) = I(\mathsf{p})$, where $\mathsf{p} \in {\cal A}$

\item $\lsem (\phi_1 \wedge \phi_2) \rsem (I) = \min\{\lsem \phi_1 \rsem (I), \lsem \phi_2 \rsem (I)\}$

\item $\lsem (\phi_1 \vee \phi_2) \rsem (I) = \max\{\lsem \phi_1 \rsem (I), \lsem \phi_2 \rsem (I)\}$

\item $\lsem (\neg \phi)\rsem (I) = \overline{\lsem \phi \rsem (I)}$

\item $\lsem (\phi_1 \gg \phi_2) \rsem (I)  =  \lsem \phi_1 \rsem (I) \gg \lsem \phi_2 \rsem (I)$.

\end{itemize}
where $\min$ and $\max$ are defined with respect to the ordering of
Definition~\ref{lexicographic-ordering}.
\end{definition}
\begin{table*}[t]
\begin{minipage}{0.47\textwidth}
\begin{tabular}{ccc@{\hskip 6pt}l@{\hskip 6pt}l}
\toprule
$x$   & $y$   &  $z$  &   $(x \gg y) \gg z$       & $x \gg (y \gg z)$ \\ 
\midrule
$[F]$ & $[F]$ & $[F]$ &   $[F]$                   &  $[F]$ \\ 
$[F]$ & $[F]$ & $[T]$ &   $[0,F,T]$               &  $[0,F,0,F,T]$ \\ 
$[F]$ & $[T]$ & $[F]$ &   $[0,0,F,T,F]$           &  $[0,F,0,T,F]$  \\ 
$[F]$ & $[T]$ & $[T]$ &   $[0,0,F,T,T]$           &  $[0,F,T]$  \\ 
$[T]$ & $[F]$ & $[F]$ &   $[0,0,T,F,F]$           &  $[0,T,F]$  \\ 
$[T]$ & $[F]$ & $[T]$ &   $[0,0,T,F,T]$           &  $[0,T,0,F,T]$  \\ 
$[T]$ & $[T]$ & $[F]$ &   $[0,T,F]    $           &  $[0,T,0,T,F]$ \\ 
$[T]$ & $[T]$ & $[T]$ &   $[T]        $           &  $[T]$  \\ 
\bottomrule
\end{tabular}
\vspace{0.2cm}
\captionof{table}{Evaluation of $(x \gg y) \gg z$ and $x \gg (y \gg z)$
         under different truth assignments.}
\label{expressions-table}
\end{minipage}
\qquad\qquad
\begin{minipage}{0.4\textwidth}
\begin{tabular}{l@{\hskip 12pt}l}
\toprule
Expression               & List           \\
\midrule
$([F] \gg [F]) \gg [F]$  & $[F]$          \\
$[F] \gg ([F] \gg [F])$  & $[F]$          \\
$[F] \gg ([F] \gg [T])$  & $[0,F,0,F,T]$  \\
$[F] \gg ([T] \gg [F])$  & $[0,F,0,T,F]$  \\
$[F] \gg ([T] \gg [T])$  & $[0,F,T]$      \\
$([F]\gg [F]) \gg [T] $  & $[0,F,T]$      \\
$([F] \gg [T]) \gg [F]$  & $[0,0,F,T,F]$  \\
$([F] \gg [T]) \gg [T]$  & $[0,0,F,T,T]$  \\
\midrule
$([T] \gg [F]) \gg [F]$  & $[0,0,T,F,F]$  \\
$([T] \gg [F]) \gg [T]$  & $[0,0,T,F,T]$  \\
$([T] \gg [T]) \gg [F]$  & $[0,T,F]$      \\
$[T] \gg ([F] \gg [F])$  & $[0,T,F]$      \\
$[T] \gg ([F] \gg [T])$  & $[0,T,0,F,T]$  \\
$[T] \gg ([T] \gg [F])$  & $[0,T,0,T,F]$  \\
$[T] \gg ([T] \gg [T])$  & $[T]$          \\
$([T] \gg [T]) \gg [T]$  & $[T]$          \\
\bottomrule
\end{tabular}
\vspace{0.2cm}
\captionof{table}{Ordered values for the expressions of Table~\ref{expressions-table}.}
\label{ranking-table}
\end{minipage}
\vspace{-0.7cm}
\end{table*}
Using the above semantics, Table~\ref{expressions-table} presents the possible
values that the expressions $(x \gg y) \gg z$ and $x \gg (y \gg z)$ can take
when the propositional atoms $x,y,z$ receive the standard truth values $[F]$ and
$[T]$.
Table~\ref{ranking-table} presents the same values as that of
Table~\ref{expressions-table} ordered from the lowest to the highest. The table
is divided in two layers. The upper layer (first 8 entries) contains values that
correspond to different degrees of false (ie., that have a $\mathit{sign}$ equal
to $F$), while the lower layer (next 8 entries) correspond to different degrees
of true (have a $\mathit{sign}$ equal to $T$).
Given a formula $\phi$ and a set of different truth assignments, we can find the
most preferable truth assignments for $\phi$ by calculating the meaning of
$\phi$ under all these assignments, and comparing the results.
\begin{definition}
Let $\phi$ be a formula and consider truth assignments $I_1,I_2$. We will say
that $I_2$ is {\em preferable to $I_1$ with respect to formula $\phi$} if $\lsem
\phi \rsem (I_1) < \lsem \phi \rsem (I_2)$.
\end{definition}
\begin{example}
Let $\phi$ be the formula ${\tt electric}\gg {\tt (fast}\gg {\tt blue)}$
of Example~\ref{car-example}. Consider the truth assignments:
\[
\begin{array}{lll}
I_1 & = & \{({\tt electric},[T]),({\tt fast},[T]),({\tt blue},[T])\}\\
I_2 & = & \{({\tt electric},[T]),({\tt fast},[T]),({\tt blue},[F])\}\\
I_3 & = & \{({\tt electric},[F]),({\tt fast},[T]),({\tt blue},[T])\}
\end{array}
\]
Using Table~\ref{expressions-table} we get that $I_1(\phi) = [T]$,
$I_2(\phi) = [0,T,0,T,F]$ and $I_3(\phi) = [0,F,T]$. By lexicographically
comparing the corresponding lists, we get that with respect to formula $\phi$,
the truth assignment $I_1$ is preferable to $I_2$ which is preferable to $I_3$.
\end{example}
\begin{example}
Let $\phi$ be the formula:
$${\tt (electric}\oplus {\tt diesel)} \gg {\tt (fast}\wedge \neg{\tt expensive)}$$
where $\oplus$ is the usual exclusive-or operation. One can easily verify that
both of the interpretations:
\[
\begin{array}{lll}
I_1 & = & \{({\tt electric},[T]),({\tt diesel},[F]),({\tt fast},[T]),({\tt expensive},[F])\}\\
I_2 & = & \{({\tt electric},[F]),({\tt diesel},[T]),({\tt fast},[T]),({\tt expensive},[F])\}
\end{array}
\]
are maximally preferable because they both assign to $\phi$ the truth
value $[T]$.
\end{example}
One could define a consequence relation for lexicographic logic based on
maximally preferred truth assignments (see, for example, the corresponding such
notion for QCL in~\cite[Definition 6]{BrewkaBB04}). Due to space restrictions,
we do not pursue this direction in the present paper, apart from a slightly more
extended discussion at the end of Section~\ref{related-work}.

\section{Some Properties of Lexicographic Logic}\label{properties}
One key property that needs to be established for lexicographic logic, is that
the ``$\gg$'' operator {\em indeed} implements lexicographic priority.
Intuitively speaking, this means that when we apply ``$\gg$'' on two distinct
pairs of arguments, and the first pair is lexicographically smaller than the
second one, then the first result is lexicographically smaller than the second
one. More formally, we define the lexicographic ordering on pairs, as follows:
\begin{definition}
For all $u_1,u_2,v_1,v_2 \in \mathbb{V}$, we write $(u_1,u_2) <_{L} (v_1,v_2)$
if either $u_1 < v_1$, or $u_1 = v_1$ and $u_2 < v_2$.
\end{definition}
The above property is expressed by the following theorem:
\begin{theorem}\label{lexicographic-theorem}
Let $u_1,u_2,v_1,v_2 \in \mathbb{V}$. Then, $(u_1,u_2) <_{L} (v_1,v_2)$ iff
$(u_1 \gg u_2) < (v_1 \gg v_2)$.
\end{theorem}
\ifincludeproof
\begin{proof}
Consider first the left to right implication. We examine cases with respect to
$(u_1,u_2)$. If $(u_1,u_2) = ([F],[F])$ then $(v_1,v_2) \neq ([F],[F])$. But
then, $(u_1 \gg u_2) = [F] <  (v_1 \gg v_2)$, because the first element of $(v_1
\gg v_2)$ will either be $0$ or $T$. Notice also that it can not be $(u_1,u_2) =
([T],[T])$, because in this case there do not exist $v_1,v_2$ such that
$(u_1,u_2) <_{L} (v_1,v_2)$. If $(u_1,u_2) \neq ([F],[F])$ and $(u_1,u_2) \neq
([T],[T])$, then, by the definition of $\gg$, the first element of $(u_1 \gg
u_2)$ will be $0$. If $(v_1,v_2) = ([T],[T])$ then $(v_1 \gg v_2)=[T]$ and $(u_1
\gg u_2) < (v_1 \gg v_2)$ obviously holds. Assume that $(v_1,v_2) \neq
([T],[T])$. Then, by the definition of $\gg$, the first element of $(u_1 \gg
u_2)$ will also be $0$. Since $(u_1,u_2) <_{L} (v_1,v_2)$, it will either be
$u_1 < v_1$, or $u_1 = v_1$ and $u_2 < v_2$. If $u_1 < v_1$ then
$[0]\mdoubleplus u_1  \mdoubleplus u_2 < [0]\mdoubleplus v_1  \mdoubleplus v_2$
and therefore $(u_1 \gg u_2) < (v_1 \gg v_2)$. If $u_1 = v_1$ and $u_2 < v_2$,
then again $[0]\mdoubleplus u_1  \mdoubleplus u_2 < [0]\mdoubleplus v_1
\mdoubleplus v_2$ and therefore $(u_1 \gg u_2) < (v_1 \gg v_2)$.

Consider now the right to left implication. If $(u_1,u_2) = ([F],[F])$ then
$u_1\gg u_2 = [F]$ and it has to be $(v_1\gg v_2) \neq [F]$, which implies that
$(v_1,v_2) \neq ([F],[F])$ and therefore $(u_1,u_2) <_L (v_1,v_2)$. If
$(u_1,u_2) = ([T],[T])$ then there can not exist $v_1,v_2$ such that
$[T]<(v_1\gg v_2)$. Consider now the case $(u_1,u_2) \neq ([F],[F])$ and
$(u_1,u_2) \neq ([T],[T])$. Then, $(u_1\gg u_2) = [0]\mdoubleplus u_1
\mdoubleplus u_2$. If $(v_1,v_2)=([T],[T])$, then, obviously, $(u_1,u_2) <_L
(v_1,v_2)$. Otherwise, $(v_1\gg v_2) = [0]\mdoubleplus v_1 \mdoubleplus v_2$.
Let $u= u_1 \mdoubleplus u_2$ and $v = v_1 \mdoubleplus v_2$, and let $u[i]$ and
$v[i]$ denote the $i$'th element of $u$ and $v$ respectively. Since $(u_1 \gg
u_2) < (v_1 \gg v_2)$, there exists $i$ such that $u[i] < v[i]$. If $i \leq
|u_1|$ and $i\leq |v_1|$, then $u_1 < v_1$ and therefore $(u_1,u_2) <_L
(v_1,v_2)$. The case $i \leq |u_1|$ and $i > |v_1|$ is not applicable because
then $v_1$ would be a proper prefix of $u_1$, which is impossible from
Lemma~\ref{prefix_lemma}. Similarly if $i > |u_1|$ and $i \leq |v_1|$. Finally,
if $i>|u_1|$ and $i> |v_1|$, then $u_1 = v_1$. Since $(u_1 \gg u_2) < (v_1 \gg
v_2)$, we get $u_2 < v_2$, which implies that $(u_1,u_2) <_L (v_1,v_2)$.
\end{proof}
\fi
Two formulas $\phi_1$ and $\phi_2$ are semantically equivalent (denoted by $\phi_1 \equiv
\phi_2$) iff for every truth assignment $I$, $\lsem \phi_1\rsem (I) = \lsem
\phi_2\rsem (I)$. One basic property that lexicographic logic inherits from
propositional logic, is the {\em substitutivity} of logically equivalent
formulas (see, for example, \cite[pp. 20-21]{Fitting96}). This property holds
due to the compositional semantics of lexicographic logic
(Definition~\ref{semantics}), and can be established by structural induction.
\begin{lemma}
Let $\phi$ be a formula of lexicographic logic, $\psi$ be a subformula of $\phi$
and $\psi'$ be a formula such that $\psi \equiv \psi'$. Then, $\phi \equiv
\phi[\psi \leftarrow \psi']$, where $\phi[\psi \leftarrow \psi']$ is the formula
that results from $\phi$ by replacing the subformula $\psi$ with $\psi'$.
\end{lemma}

The following are some basic properties of lexicographic logic that can be
easily established:
\begin{lemma}
For all formulas $\phi_1,\phi_2,\phi_3, \phi$, the following equivalences hold:
\begin{itemize}
\item $(\phi_1 \vee \phi_2) \gg \phi_3  \equiv (\phi_1 \gg \phi_3) \vee (\phi_2 \gg \phi_3)$

\item $\phi_1 \gg (\phi_2 \vee \phi_3)  \equiv (\phi_1 \gg \phi_2) \vee (\phi_1 \gg \phi_3)$

\item $(\phi_1 \wedge \phi_2) \gg \phi_3  \equiv (\phi_1 \gg \phi_3) \wedge (\phi_2 \gg \phi_3)$

\item $\phi_1 \gg (\phi_2 \wedge \phi_3) \equiv (\phi_1 \gg \phi_2) \wedge (\phi_1 \gg \phi_3)$

\item $\neg (\phi_1 \gg \phi_2) \equiv (\neg \phi_1) \gg (\neg  \phi_2)$

\item $\neg(\neg \phi) \equiv \phi$

\end{itemize}
\end{lemma}
\ifincludeproof
\begin{proof}
We prove the first statement. The remaining ones are established in a similar
way. Let $I$ be an arbitrary truth assignment. We show that:
$$\lsem (\phi_1 \vee \phi_2) \gg \phi_3 \rsem (I) = \lsem (\phi_1 \gg \phi_3)
\vee (\phi_2 \gg \phi_3)\rsem (I)$$
Using Definition~\ref{semantics}, the left hand side evaluates to:
$$\max\{\lsem \phi_1 \rsem (I), \lsem \phi_2\rsem (I)\} \gg \lsem \phi_3
\rsem (I)$$
while the right hand side evaluates to:
$$\max\{\lsem \phi_1\rsem (I) \gg \lsem \phi_3 \rsem (I), \lsem
\phi_2\rsem (I) \gg \lsem \phi_3\rsem (I)\}$$
If $\lsem \phi_1 \rsem (I) = \lsem \phi_2\rsem (I)$, then the equality of the
above two statements obviously holds. Otherwise, assume without loss of
generality that $\lsem \phi_1 \rsem (I) > \lsem \phi_2\rsem (I)$. By
Theorem~\ref{lexicographic-theorem}, it is:
$$(\lsem \phi_1\rsem (I) \gg \lsem \phi_3 \rsem (I)) > (\lsem \phi_2\rsem (I)
\gg \lsem \phi_3\rsem (I))$$
and the equality of the two statements again holds.
\end{proof}
\fi

\section{Numerical Representation}
\label{numerical}
In this section, we demonstrate that the elements of $\mathbb{V}$ can be mapped
to rational numbers so as that their ordering is preserved. More formally, we
demonstrate that there exists a function $\mathit{val}: \mathbb{V} \rightarrow
\mathbb{Q}$ such that for all $u,v \in \mathbb{V}$, if $u < v$ then
$\mathit{val}(u) < \mathit{val}(v)$. The key idea of defining $\mathit{val}$
is that the elements of $\mathbb{V}$ can
be considered as numbers in the interval $[-1,1]$ written in the {\em balanced
ternary number system}~\cite[p. 207]{knuth1998art2}. As Donald Knuth poses it,
balanced ternary is ``{\em perhaps the prettiest number system of all}''.
Balanced ternary is a ternary number system in which the coefficients are the
numbers $-1$, $0$, and $1$ (instead of $0$, $1$, and $2$, as it happens in
standard ternary notation). It  is called ``balanced ternary'' because the
coefficients are symmetrical around zero.

We consider the elements of $\mathbb{V}$ as representing balanced ternary
numbers in the interval $[-1,1]$. More specifically, $\mathit{val}(F)=-1$,
$\mathit{val}(T)=1$, $\mathit{val}(0)=0$, and for every other element
$u=[u_1,\ldots,u_{n}] \in \mathbb{V}$, we derive a rational number in the
interval $(-1,1)$ by viewing $u$ as a balanced ternary number. Since we want the
value of the number to belong in  the interval $(-1,1)$, we calculate its value
by using the powers of $\frac{1}{3}$ (in the same way that the value of the
fractional part of a decimal number can be calculated using the powers of
$\frac{1}{10}$). Formally:
\[
      \mathit{val}([u_1,\ldots,u_{n}]) = \sum_{i=1}^{n}\mathit{val}(u_i)\cdot \frac{1}{3^{i-1}}
\]
Table~\ref{value-table} depicts the numerical values that correspond to all
elements of $\mathbb{V}$ of size less than or equal to $5$.
\begin{table}
\vspace{-0.7cm}
\begin{center}
\begin{tabular}{lc}
\toprule
 $v \in \mathbb{V}$    & $\mathit{val}(v)$    \\
\midrule
 $[F]$                 &  $-1$              \\
 $[0,F,0,F,T]$         &  $\sfrac{-29}{81}$ \\
 $[0,F,0,T,F]$         &  $\sfrac{-25}{81}$ \\
 $[0,F,T]$             &  $\sfrac{-2}{9}$  \\
 $[0,0,F,T,F]$         &  $\sfrac{-7}{81}$ \\
 $[0,0,F,T,T]$         &  $\sfrac{-5}{81}$ \\
 $[0,0,T,F,F]$         &  $\sfrac{\quad\; 5}{81}$ \\
 $[0,0,T,F,T]$         &  $\sfrac{\quad\; 7}{81}$ \\
 $[0,T,F]$             &  $\sfrac{\quad\; 2}{9}$  \\
 $[0,T,0,F,T]$         &  $\sfrac{\quad\; 25}{81}$ \\
 $[0,T,0,T,F]$         &  $\sfrac{\quad\; 29}{81}$ \\
 $[T]$                 &  $\quad\; 1$  \\
\bottomrule
\end{tabular}
\end{center}
\caption{Rational values for elements of $\mathbb{V}$.}\label{value-table}
\vspace{-0.7cm}
\end{table}
The following lemma justifies why we can use the numerical representation of
the elements of $\mathbb{V}$ as an equivalent alternative:
\begin{lemma}
For all $u,v \in \mathbb{V}$, if $u < v$, then $\mathit{val}(u) < \mathit{val}(v)$.
\end{lemma}
\ifincludeproof
\begin{proof}
If $u=[F]$ and $v=[T]$ the result obviously holds. If $u=[F]$ and $v\neq [T]$
then $v=[v_1,\ldots,v_m]$, where $v_1=0$. In the extreme case where for all $i
\geq 2$, $v_i=F$, we get:
$$\mathit{val}(v) = \sum_{i=2}^{m}\mathit{val}(v_i)\cdot \frac{1}{3^{i-1}} \geq -\frac{1}{2}$$
Therefore, in this case $\mathit{val}(u) < \mathit{val}(v)$.

Consider now the case where $u=[u_1,\ldots,u_m]$ and $v=[v_1,\ldots,v_n]$, with
$u_1=v_1=0$. Since $u<v$, there exists $k$ with $2 \leq k \leq
\min\{m,n\}$ such that for all $i<k$, $u_i = v_i$, and $u_k < v_k$. We
show that $\mathit{val}(u) < \mathit{val}(v)$, or equivalently that
$\mathit{val}(u) - \mathit{val}(v) < 0$. We have:
$$\mathit{val}(u) - \mathit{val}(v) = \sum_{i=2}^n \mathit{val}(u_i)\cdot
\frac{1}{3^{i-1}} -\sum_{i=2}^m \mathit{val}(v_i)\cdot \frac{1}{3^{i-1}}$$
Notice now that $\mathit{val}(u_k)\cdot \frac{1}{3^{k-1}} -
\mathit{val}(v_k)\cdot \frac{1}{3^{k-1}} \leq -\frac{1}{3^{k-1}}$ (when $u_k = 0$ and $v_k = T$,
or when $u_k=F$ and $v_k = 0$). We demonstrate that no matter what the remaining
digits of $u$ and $v$ are, they can not ``close the gap'' of
$-\frac{1}{3^{k-1}}$. Take the extreme case where all the $u_i$, $i>k$, are
equal to $T$ and all the $v_i$, $i>k$, are equal to $F$. Then:
\[
\begin{array}{lll}
& & \sum_{i=k+1}^n \mathit{val}(u_i)\cdot \frac{1}{3^{i-1}} -\sum_{i=k+1}^m \mathit{val}(v_i)\cdot \frac{1}{3^{i-1}}\\\\
&=& \sum_{i=k+1}^{n} \frac{1}{3^{i-1}} +\sum_{i=k+1}^m \frac{1}{3^{i-1}}\\\\
&<& \sum_{i=k+1}^{\infty} \frac{1}{3^{i-1}} +\sum_{i=k+1}^{\infty} \frac{1}{3^{i-1}}\\\\
&=& 2 \cdot \sum_{i=k+1}^{\infty} \frac{1}{3^{i-1}}\\\\
&=& \frac{1}{3^{k-1}}
\end{array}
\]
Thus, the gap can not close and $\mathit{val}(u) - \mathit{val}(v) < 0$.
\end{proof}
\fi

The above discussion suggests that the semantics of lexicographic logic can be
equivalently expressed using a special subset of the set $\mathbb{Q}$ of
rational numbers. In a potential implementation of a query system based on
lexicographic logic, numerical values that rank query results would convey a
much better intuition than the lists of truth values.

\section{Modeling Alternative Preferential Operators}\label{alternative}
In this section, we demonstrate that we can use ``$\gg$'' as a primitive operator in
order to define other interesting connectives that express levels of preference.
We start with two well-known such operators~\cite{DuboisP13}, namely ``{\em and if possible}''
and ``{\em or at least''}. We will denote the former by ``$\&$'' and the latter by
``$\times$''.

\subsection{Modeling the ``and if possible'' operator}\label{the-if-possible}
The intuitive meaning of $(x\, \& \, y)$ is ``{\em I want $x$ and if possible additionally $y$}''. 
The behaviour we expect from ``$\&$'' when applied to {\em classical} truth values,
is depicted in Table~\ref{truth-table-if-possible}.
\begin{table}
\mbox{}\hfill
\begin{minipage}[t]{0.4\linewidth}\centering
\begin{center}
\begin{tabular}{ccl}
\toprule
$x$   & $y$    & $x \, \& \, y$ \\
\midrule
$[F]$ & $[F]$   & $[F]$\\
$[F]$ & $[T]$   & $[F]$\\
$[T]$ & $[F]$   & $[T]\gg[F]$ \\
$[T]$ & $[T]$   & $[T]$\\
\bottomrule
\end{tabular}
\end{center}
\caption{The truth table of ``$\&$'' for standard truth values.}
\label{truth-table-if-possible}
\end{minipage}
\hfill
\begin{minipage}[t]{0.4\linewidth}\centering
\begin{center}
\begin{tabular}{ccl}
\toprule
$x$   & $y$    & $x \, \times \, y$ \\
\midrule
$[F]$ & $[F]$   & $[F]$ \\
$[F]$ & $[T]$   & $[T] \gg [F]$ \\
$[T]$ & $[F]$   & $[T]$ \\
$[T]$ & $[T]$   & $[T]$ \\
\bottomrule
\end{tabular}
\end{center}
\caption{The truth table of ``$\times$'' for standard truth values.}
\label{truth-table-or-failing-that}
\end{minipage}
\hfill\mbox{}
\end{table}
The intuitive reading of the table is as follows: if the first argument of
``$\&$'' is false, then the result is false (no matter what the second argument
is). If both arguments are true, then the result is true. If, however, the first
argument is true and the second is false, then we are partially satisfied; this
is expressed by the value $[T]\gg [F]$ which is less than the absolute true
value $[T]$. It turns out that ``$\&$'' can be described as a derived operator using
``$\gg$''. Formally:
\[
      x\, \& \, y = x \gg (x \wedge y)
\]
It can easily be verified that the above definition produces the values
of Table~\ref{truth-table-if-possible}. Notice also that the above definition is
meaningful even if ``$\&$'' is applied on non-classical elements of
$\mathbb{V}$, ie., elements that are different from $[F]$ or $[T]$. In this
case, the intuition of the above definition is that ``$\&$'' puts a strong
emphasis on the value of its first argument ($x$ in our case). In particular, if
$x$ has a false value (of whatever degree), then even if $y$ has a true value,
this value will be degraded (through the $\wedge$ operation).

We can easily verify that our definition satisfies an intuitive equivalence
observed in~\cite{DuboisP13}, namely that ``$x$ and if possible $y$'' is
equivalent to ``$x$ and if possible $x \wedge y$''. Indeed:
\[
\begin{array}{lll}
x\, \& \, (x \wedge y) \, = \, x \gg (x \wedge (x \wedge y)) \, = \, x \gg (x \wedge y) \, = \, x \, \& \,y
\end{array}
\]
As we are going to see in the next subsection, our definition of ``$\&$''
satisfies some intuitive and expected properties with respect to the ``or at
least'' operator.

\subsection{Modeling the ``or at least'' operator}\label{the-times}
The intuitive meaning of $(x\, \times \, y)$ is ``{\em I want $x$, or failing
that, $y$}''. The behaviour we expect from ``$\times$'' when applied to
{\em classical} truth values, is depicted in Table~\ref{truth-table-or-failing-that}.
The intuitive reading of the table is as follows: if both arguments of
``$\times$'' are false, then the result is false. If the first argument is true,
then the result is true (no matter what the second argument is). If, however,
the first argument is false and the second is true, then we are still satisfied,
but not as much as if the first argument was true; this is expressed by the
value $[T]\gg [F]$ which is not absolutely true (but is still true). We can 
express ``$\times$'' using ``$\gg$'', as follows:
\[
      x\, \times \, y = (x \vee y) \gg x
\]
The above definition is meaningful even for values of $\mathbb{V}$
that are non-classical. If either $x$ or $y$ have a true value (of whatever form), 
then the result will also be a true value of some form. In particular, if $x$ has 
some false value, we degrade the overall result because we are not entirely 
satisfied (this is achieved by the occurrence of $x$ in the right of ``$\gg$'').

Using our definition of ``$\times$'', we can easily verify that, as observed
in~\cite{DuboisP13}, ``$x$ or at least $y$'' is equivalent to ``$x$ or at least
$x \vee y$''. Indeed:
\[
\begin{array}{lll}
x\, \times\, (x \vee y) \, = \, (x \vee (x \vee y)) \gg x \, = \, (x \vee y) \gg x \, = \, x \, \times \,y
\end{array}
\]
Notice also that there is a connection between the ``$\times$'' and ``$\&$''
operators, observed in~\cite{DuboisP13}, which the following lemma demonstrates.
\begin{lemma}
For all $x,y\in \mathbb{V}$, it holds:
\begin{itemize}
\item $x \times y = (x \vee y)\, \& \, x$
\item $x \, \& \, y = (x \wedge y) \times x$.
\end{itemize}
\end{lemma}
\ifincludeproof
\begin{proof}
We have:
\[
\begin{array}{lll}
(x \vee y)\,\&\, x  & = & (x \vee y) \gg ((x \vee y) \wedge x) \\
                    & = & (x \vee y) \gg x \\
                    & = & x \times y
\end{array}
\]
Also:
\[
\begin{array}{lll}
(x \wedge y) \times x  & = & ((x \wedge y) \vee x) \gg (x \wedge y)\\
                       & = & x \gg (x \wedge y)\\
                       & = & x \, \& \, y
\end{array}
\]
This completes the proof of the lemma.
\end{proof}
\fi

\subsection{Modeling Hierarchical Operators}

As it turns out, we can use ``$\gg$'' to model preference operators that are
``hierarchical'' in nature. Consider, for example, the ``$x$ or at least $y$''
operator and assume for simplicity that $x$ and $y$ are standard truth values,
namely $[F]$ and $[T]$. The output of this operator can belong to three distinct
levels of a hierarchy: the top-level value is attained when the input values are
$x=[T]$ and $y=[T]$, or when $x=[T]$ and $y=[F]$; the next level is attained by
$x=[F]$ and $y=[T]$; finally, for $x=[F]$ and $y=[F]$ we obtain the lowest
value. A similar hierarchy exists in the definition of the ``$\&$'' operator.

The above ideas can be generalized as follows. Assume we want to construct an
operator $f$ on $n$ input arguments $x_1,\ldots,x_n$. Moreover, assume that we
want the value of $f(x_1,\ldots,x_n)$ to belong to $m \leq 2^n$ distinct levels
of a hierarchy. We mark all those inputs for which we want $f$ to attain the
largest possible value. For all those inputs, we define $f$ to return the top
possible truth value, namely $[T]$. We then mark all those inputs for which we
want $f$ to return the immediately lower truth value. For all those inputs, we
define $f$ to return a truth value below $[T]$, for example $[T]\gg \cdots \gg
[T]\gg [F]$ (where ``$\gg$'' appears $n-1$ times). The next level will receive
an even smaller value, and so on. We continue this process until all levels of
the hierarchy are exhausted. Notice that the choice of the exact truth value
that we assign to each level, is not important, as long as ``a more important
level'' receives a higher truth value than the ``less important'' ones. This
process helps us create a truth table, as illustrated by the following example.
\begin{example}
Assume we want to construct a ternary operator $\mathit{more}(x_1,x_2,x_3)$,
which is interpreted as ``{\em the more the better}''. We would like
$\mathit{more}(x_1,x_2,x_3)$ to attain its highest value when all of its inputs
are $[T]$. Additionally, we want $\mathit{more}(x_1,x_2,x_3)$ to obtain a lower
truth value when exactly two of its arguments are $[T]$. Finally, when exactly
one of the arguments is equal to $[T]$, we want $\mathit{more}(x_1,x_2,x_3)$ to
return an even lower truth value. We create the truth table shown in
Table~\ref{more-table}. For symmetry reasons we represent $[T]$ by $[T]\gg
[T]\gg [T]$. The second truth value that we use is $[T]\gg [T] \gg [F]$ and the
third one $[T]\gg [F] \gg [F]$.
\begin{table}
\vspace{-0.7cm}
\begin{center}
\begin{tabular}{llll}
\toprule
$x_1$   & $x_2$   &  $x_3$  &   $\mathit{more}(x_1,x_2,x_3)$ \\
\midrule
$[F]$ & $[F]$  & $[F]$ &   $[F]\gg [F]\gg [F]$ \\
$[F]$ & $[F]$  & $[T]$ &   $[T]\gg [F]\gg [F]$ \\
$[F]$ & $[T]$  & $[F]$ &   $[T]\gg [F]\gg [F]$ \\
$[F]$ & $[T]$  & $[T]$ &   $[T]\gg [T]\gg [F]$ \\
$[T]$ & $[F]$  & $[F]$ &   $[T]\gg [F]\gg [F]$ \\
$[T]$ & $[F]$  & $[T]$ &   $[T]\gg [T]\gg [F]$ \\
$[T]$ & $[T]$  & $[F]$ &   $[T]\gg [T]\gg [F]$ \\
$[T]$ & $[T]$  & $[T]$ &   $[T]\gg [T]\gg [T]$ \\
\bottomrule
\end{tabular}
\end{center}
\caption{Truth table for $\mathit{more}$.}
\label{more-table}
\vspace{-0.7cm}
\end{table}
\end{example}
Let us call the last column of the truth table (ie., the one that contains the
desired truth values for the function $f$), the {\em assignment column}. We create
the function definition:
\[
      f(x_1,\ldots,x_n) = E_1 \gg \cdots \gg E_n
\]
where the $E_i$ are standard propositional formulas (ie., they do not contain
the ``$\gg$'' operator) constructed using the variables $x_1,\ldots,x_n$. Each
$E_i$ is constructed as the disjunction of a set of conjunctive terms; each such
conjunctive term corresponds to a row of the truth table which has in the $i$'th
position of the assignment column the value $[T]$.
\begin{example}
We construct a function $f(x_1,x_2,x_3) = E_1\gg E_2 \gg E_3$ for the truth
table given in Table~\ref{more-table}. First we construct the disjunction of
terms for $E_1$. We notice that the rows 2-8 of Table~\ref{more-table} have the
value $[T]$ in the first position of the assignment column. Therefore, $E_1$ is
equal to:
\[
\begin{array}{l}
(\overline{x_1}\wedge \overline{x_2} \wedge x_3) \vee (\overline{x_1}\wedge x_2 \wedge \overline{x_3}) \vee (\overline{x_1}\wedge x_2 \wedge x_3) \vee \\
(x_1\wedge \overline{x_2} \wedge \overline{x_3}) \vee (x_1\wedge \overline{x_2} \wedge x_3) \vee (x_1\wedge x_2 \wedge \overline{x_3}) \vee (x_1\wedge x_2 \wedge x_3)
\end{array}
\]
The above easily simplifies to $E_1 = x_1 \vee x_2 \vee x_3$.
To construct the expression for $E_2$ we observe that the rows 4, 6, 7 and 8 of
Table~\ref{more-table} have the value $[T]$ in the second position of the
assignment column. Therefore, $E_2$ is equal to:
\[
\begin{array}{l}
(\overline{x_1}\wedge x_2 \wedge x_3) \vee (x_1\wedge \overline{x_2} \wedge x_3) \vee (x_1\wedge x_2 \wedge \overline{x_3}) \vee (x_1\wedge x_2 \wedge x_3)
\end{array}
\]
This simplifies to $E_2 = (x_1 \wedge x_2) \vee (x_2 \wedge x_3) \vee (x_1 \wedge x_3)$.
Finally, to construct the expression for $E_3$ we observe that only the 8th
row of Table~\ref{more-table} has the value $[T]$ in the third position of the
assignment column. Therefore, $E_3 = (x_1 \wedge x_2 \wedge x_3)$.
Overall, the expression for $\mathit{more}(x_1,x_2,x_3)$ is:
\[
\begin{array}{l}
 (x_1 \vee x_2 \vee x_3) \gg
 ((x_1 \wedge x_2) \vee (x_2 \wedge x_3) \vee (x_1 \wedge x_3)) \gg
 (x_1 \wedge x_2 \wedge x_3)
\end{array}
\]
One can verify that the above expression produces the values
of Table~\ref{more-table}.
\end{example}
It is worth noting that the definitions of ``$\&$'' and ``$\times$'' given in
Subsections~\ref{the-if-possible} and~\ref{the-times} respectively, were both derived
using the systematic procedure presented in the subsection.

\section{Related Work}\label{related-work}
There exists a great variety of preference representation formalisms that have
been developed mainly in the areas of artificial intelligence, databases, and
logic programming. Nice surveys that outline the main approaches in these three
areas are~\cite{DomshlakHKP11},~\cite{StefanidisKP11}, and~\cite{SakamaI00},
respectively. In general, preference representation formalisms are
classified~\cite{StefanidisKP11} as either ``{\em quantitative} or ``{\em
qualitative}''. In the quantitative approach, numerical values are used in the
syntax of the preference specification language in order to express degrees
of preference. On the other hand, in the qualitative approach, preferences are
expressed by implicitly establishing a preference ordering relation between the
objects under consideration. Lexicographic logic falls somewhere in between: its
syntax is qualitative because preferences are expressed implicitly through the
``$\gg$'' operator; however, its semantics is quantitative because, as shown in
Section~\ref{numerical}, formulas essentially receive rational number values
when evaluated.

The work proposed in this paper suggests the use of a high-level logic syntax in
order to represent preferences. The research works that we are aware of and that
are closer to our contribution, are the following:
\begin{itemize}
\item The use of {\em negation-as-failure} in logic programming which
      acts as a quantitative preferential operator.

\item The {\em Qualitative Choice Logic (QCL)} approach proposed in~\cite{BrewkaBB04}
      and extended in the context of logic programming in~\cite{BrewkaNS04}.

\item The {\em flexible queries} approach described in~\cite{DuboisP13} that is
      based on possibilistic logic~\cite{DuboisP14}.

\item The {\em infinite-valued} approach proposed in~\cite{Aga05,AW05} and extended in
      the context of logic programming in~\cite{RondogiannisT15}.
\end{itemize}
In the following, we give a brief comparison of our approach with each one of the
above formalisms.

\vspace{0.15cm}
\noindent
{\bf Negation-as-failure}:
As remarked in~\cite{SakamaI00}, stratified logic programs~\cite{AptBW88}
introduce a form of preferences in logic programming. A very concise discussion
of this issue can also be found in~\cite{Przymusinska90semanticissues} where it
is argued that in a logic program with negation, the negative literals in the
body of a rule have a {\em higher priority} for falsification in the intended
meaning of the program than the atom in the head of the same rule. In other
words, the well-known ``{\em not}'' operator of logic programming is an implicit
preferential operator that plays a role similar to that of the ``$\gg$''
operator in lexicographic logic. Of course, there exists a significant
difference between the two operators: negation-as-failure is used to
express a very specific type of preference related to the falsification of
atoms, while ``$\gg$'' can be used to express more general types of preferences.

\vspace{0.15cm}
\noindent
{\bf Qualitative Choice Logic (QCL)}:
In~\cite{BrewkaBB04}, propositional logic is extended with the new connective
``$\times$''. Intuitively, $A \times B$ is read ``{\em if possible $A$, but if
$A$ is impossible then at least $B$''}.  QCL has a similar philosophy as
lexicographic logic in the sense that preferences are represented implicitly
using a new operator. However, there are important differences. QCL is built on
the classical boolean truth domain while lexicographic logic uses a many-valued
one. It has been remarked that the semantics of QCL lead to certain
limitations~\cite{BenferhatS07}. Moreover, some intuitive tautologies, such as
$\neg (\neg \phi) \equiv \phi$ do not hold, and the notion of logical
equivalence between QCL formulas is defined in a non-standard way
(see~\cite{BrewkaBB04}[page 209]). Additionally, the semantics of QCL can be
used to prove that the operator ``$\times$'' of QCL is associative, something that
does not hold for the operator ``$\times$'' that we defined in Subsection~\ref{the-times}.
It is also worth noting that QCL is a non-monotonic logic, while Lexicographic Logic
(in its present form) has not been extended with a non-monotonic consequence relation;
a more general remark regarding this issue is given at the end of this section. Finally,
as shown in Section~\ref{alternative}, lexicographic logic can be used to define
a variety of operators, something which does not appear to be the case for QCL.

\vspace{0.15cm}
\noindent
{\bf Flexible Queries}:
The operators ``$A$ and if possible $B$'' and ``$A$ or at least $B$'' have
been studied as different forms of bipolar constraints for flexible querying
in~\cite{DuboisP13}. The semantics of these operators has been modelled
using possibilistic logic~\cite{DuboisP13,DuboisP14}. It is however unclear whether
a lexicographic priority operator like ``$\gg$'' can be encoded in the framework
of~\cite{DuboisP13} and whether arbitrary nestings of such operators can be supported.
Moreover, it is not obvious whether alternative hierarchical preferential operators
can be described. Overall, we believe that Lexicographic Logic provides a simpler
and more natural means for encoding such operators.

\vspace{0.15cm}
\noindent
{\bf Infinite-valued Logic}:
In~\cite{AW05,Aga05} a propositional query language is developed that supports
two operators which can express preferences of the form ``{\em $A$ and
optionally $B$}'' and ``{\em $A$ or alternatively $B$''}. The semantics of this
query language is based on the infinite-valued logic introduced in~\cite{RondogiannisW05}.
The language was subsequently extended with recursion obtaining the logic programming language
$\mathsf{PrefLog}$~\cite{RondogiannisT15}. The infinite-valued approach shares a
similar philosophy with lexicographic logic: preferences are expressed
implicitly using operators in the context of many-valued logics. However, there
is a significant difference between our approach and the one developed
in~\cite{Aga05,AW05}: as it was demonstrated
in~\cite{papadimitriou_master_thesis}, the infinite valued logic
of~\cite{RondogiannisW05} is not sufficient to express lexicographic preferences
(the proof uses similar arguments to the ones used in the proof of
Theorem~\ref{folk-theorem}, see for example~\cite[p. 363]{Fishburn99}). Therefore, our
present approach is more powerful than the one developed
in~\cite{Aga05,AW05} because it can express the ``optional'' and
``alternative'' preferences of the latter, and additionally it can also express
lexicographic and other hierarchical ones.

\vspace{0.15cm}
\noindent
{\bf Other Approaches}:
There are many other logical formalisms that have been defined for representing
preferences; however, to the best of our knowledge, most approaches are not
directly related to lexicographic logic. An (unavoidably) incomplete list of
such approaches can be found in~\cite{BrewkaNT08,Lang09,DomshlakHKP11}.

Many logical approaches to preference representation are {\em non-monotonic}.
Intuitively, this means that the logic provides a mechanism to distinguish among
the models of a formula and to produce only the most preferred ones. One of the
most widely used non-monotonic formalisms are {\em extended logic
programs}~\cite{GelfondL88} which have formed the basis for various preferential
formalisms~\cite{BrewkaNT08}.

Non-monotonicity is quite indispensable if we want to distinguish among the
models. However, when a logical formalism is used as a query language, we often
expect to see even models that are not necessarily optimal in terms of our preferences.
For example, when searching for hotel accommodation, we often want to also view
results that are not optimal in terms of our preferences. In its present form,
Lexicographic logic does not have a consequence relation based on maximally
preferred models. Of course, it is conceivable that it can be extended to a
non-monotonic setting but this is outside the scope of the present paper.

\section{Conclusions}\label{conclusions}
We have introduced lexicographic logic, a propositional, many-valued logic for
preference representation. Lexicographic logic has a simple semantics based on
sequences of the truth values $F$, $0$, and $T$, which can be alternatively
defined using a subset of the rational numbers in the interval $[-1,1]$.
Apart from representing lexicographic preferences (which are non-trivial to
handle from a quantitative point of view), the proposed logic can be used
to define in a simple way, alternative preferential operators.

The next step in the development of lexicographic logic, is the introduction of 
a deductive calculus for the logic.  Such an investigation could include an inference
procedure, investigation of completeness issues (with respect to the model-theoretic semantics), 
and a corresponding complexity-theoretic study. Another promising direction for 
future work would be the extension of lexicographic logic to the first-order case.
The semantics of universal quantification would most probably need to be defined as 
the least upper bound of (a possibly infinite) set of elements of $\mathbb{V}$. Since 
the least upper bound of such a set can be an infinite list, we would need to extend 
$\mathbb{V}$ to contain lists of countably infinite length; moreover, the concatenation 
operation would also have to be extended to apply to such lists. We are currently 
investigating issues such as the above ones.

\bibliography{jelia21}

\begin{thebibliography}{10}
\providecommand{\url}[1]{\texttt{#1}}
\providecommand{\urlprefix}{URL }
\providecommand{\doi}[1]{https://doi.org/#1}

\bibitem{Aga05}
Agarwal, R.: A Framework for Expressing Prioritized Constraints Using
  Infinitesimal Logic. Master's thesis, University of Victoria, Canada (2005)

\bibitem{AW05}
Agarwal, R., Wadge, W.W.: The lazy evaluation of infinitesimal logic
  expressions. In: Arabnia, H.R. (ed.) Proceedings of The 2005 International
  Conference on Programming Languages and Compilers, {PLC} 2005, Las Vegas,
  Nevada, USA, June 27-30, 2005. pp.~3--7. {CSREA} Press (2005)

\bibitem{AptBW88}
Apt, K.R., Blair, H.A., Walker, A.: Towards a theory of declarative knowledge.
  In: Minker, J. (ed.) Foundations of Deductive Databases and Logic
  Programming, pp. 89--148. Morgan Kaufmann (1988).
  \doi{10.1016/b978-0-934613-40-8.50006-3}

\bibitem{BenferhatS07}
Benferhat, S., Sedki, K.: A revised qualitative choice logic for handling
  prioritized preferences. In: Mellouli, K. (ed.) Symbolic and Quantitative
  Approaches to Reasoning with Uncertainty, 9th European Conference, {ECSQARU}
  2007, Hammamet, Tunisia, October 31 - November 2, 2007, Proceedings. Lecture
  Notes in Computer Science, vol.~4724, pp. 635--647. Springer (2007).
  \doi{10.1007/978-3-540-75256-1\_56}

\bibitem{BrewkaBB04}
Brewka, G., Benferhat, S., Berre, D.L.: Qualitative choice logic. Artificial
  Intelligence  \textbf{157}(1-2),  203--237 (2004).
  \doi{10.1016/j.artint.2004.04.006}

\bibitem{BrewkaNS04}
Brewka, G., Niemel{\"{a}}, I., Syrj{\"{a}}nen, T.: Logic programs with ordered
  disjunction. Computational Intelligence  \textbf{20}(2),  335--357 (2004).
  \doi{10.1111/j.0824-7935.2004.00241.x}

\bibitem{BrewkaNT08}
Brewka, G., Niemel{\"{a}}, I., Truszczynski, M.: Preferences and nonmonotonic
  reasoning. {AI} Magazine  \textbf{29}(4),  69--78 (2008)

\bibitem{CharalambidisRT18}
Charalambidis, A., Rondogiannis, P., Troumpoukis, A.: Higher-order logic
  programming: An expressive language for representing qualitative preferences.
  Science of Computer Programming  \textbf{155},  173--197 (2018).
  \doi{10.1016/j.scico.2017.09.002}

\bibitem{DomshlakHKP11}
Domshlak, C., H{\"{u}}llermeier, E., Kaci, S., Prade, H.: Preferences in {AI:}
  an overview. Artificial Intelligence  \textbf{175}(7-8),  1037--1052 (2011).
  \doi{10.1016/j.artint.2011.03.004}

\bibitem{DuboisP13}
Dubois, D., Prade, H.: Modeling ``and if possible'' and ``or at least'':
  Different forms of bipolarity in flexible querying. In: Pivert, O., Zadrozny,
  S. (eds.) Flexible Approaches in Data, Information and Knowledge Management,
  Studies in Computational Intelligence, vol.~497, pp. 3--19. Springer (2013).
  \doi{10.1007/978-3-319-00954-4\_1}

\bibitem{DuboisP14}
Dubois, D., Prade, H.: Possibilistic logic - an overview. In: Siekmann, J.H.
  (ed.) Computational Logic, Handbook of the History of Logic, vol.~9, pp.
  283--342. Elsevier (2014). \doi{10.1016/B978-0-444-51624-4.50007-1}

\bibitem{Fishburn99}
Fishburn, P.C.: Preference structures and their numerical representations.
  Theoretical Computer Science  \textbf{217}(2),  359--383 (1999).
  \doi{10.1016/S0304-3975(98)00277-1}

\bibitem{Fitting96}
Fitting, M.: First-Order Logic and Automated Theorem Proving, Second Edition.
  Graduate Texts in Computer Science, Springer (1996).
  \doi{10.1007/978-1-4612-2360-3}

\bibitem{GelfondL88}
Gelfond, M., Lifschitz, V.: The stable model semantics for logic programming.
  In: Kowalski, R.A., Bowen, K.A. (eds.) Logic Programming, Proceedings of the
  Fifth International Conference and Symposium, Seattle, Washington, USA,
  August 15-19, 1988 {(2} Volumes). pp. 1070--1080. {MIT} Press (1988)

\bibitem{knuth1998art2}
Knuth, D.E.: The art of computer programming, vol.~2. Addison-Wesley Longman
  Publishing Co., Boston, MA, USA, 3rd edn. (1998)

\bibitem{Lang09}
Lang, J.: Logical representation of preferences. In: Bouyssou, D., Dubois, D.,
  Pirlot, M., Prade, H. (eds.) Decision-making Process, pp. 321--363. Wiley
  (2009). \doi{10.1002/9780470611876.ch7}

\bibitem{papadimitriou_master_thesis}
Papadimitriou, G.: A Logic Query Language for Lexicographic Preferences.
  Master's thesis, National and Kapodistrian University of Athens, Greece
  (2017)

\bibitem{Przymusinska90semanticissues}
Przymusinska, H., Przymusinski, T.: Semantic issues in deductive databases and
  logic programs. In: Formal Techniques in Artificial Intelligence. pp.
  321--367. North-Holland (1990)

\bibitem{RondogiannisT15}
Rondogiannis, P., Troumpoukis, A.: Expressing preferences in logic programming
  using an infinite-valued logic. In: Falaschi, M., Albert, E. (eds.)
  Proceedings of the 17th International Symposium on Principles and Practice of
  Declarative Programming, Siena, Italy, July 14-16, 2015. pp. 208--219. {ACM}
  (2015). \doi{10.1145/2790449.2790511}

\bibitem{RondogiannisW05}
Rondogiannis, P., Wadge, W.W.: Minimum model semantics for logic programs with
  negation-as-failure. {ACM} Trans. Comput. Log.  \textbf{6}(2),  441--467
  (2005). \doi{10.1145/1055686.1055694}

\bibitem{SakamaI00}
Sakama, C., Inoue, K.: Prioritized logic programming and its application to
  commonsense reasoning. Artificial Intelligence  \textbf{123}(1-2),  185--222
  (2000). \doi{10.1016/S0004-3702(00)00054-0}

\bibitem{StefanidisKP11}
Stefanidis, K., Koutrika, G., Pitoura, E.: A survey on representation,
  composition and application of preferences in database systems. {ACM}
  Transactions on Database Systems  \textbf{36}(3),  19:1--19:45 (2011).
  \doi{10.1145/2000824.2000829}

\end{thebibliography}

\end{document}